\documentclass{article}

\usepackage[preprint]{neurips_2026}

\usepackage[utf8]{inputenc} 
\usepackage[T1]{fontenc}    
\usepackage{hyperref}       
\usepackage{url}            
\usepackage{booktabs}       
\usepackage{amsfonts}       
\usepackage{nicefrac}       
\usepackage{microtype}      
\usepackage{xcolor}         
\usepackage{graphicx}
\usepackage{mathtools}
\usepackage{array}

\newcolumntype{Y}{>{\centering\arraybackslash}X}
\usepackage{multirow}
\usepackage{multicol}
\usepackage[table,xcdraw]{xcolor}
\usepackage[normalem]{ulem}
\useunder{\uline}{\ul}{}
\usepackage{booktabs}
\usepackage{tabularx}
\usepackage{pifont}
\newcommand{\cmark}{\ding{51}} 
\newcommand{\xmark}{\ding{55}} 
\newcommand{\hb}[1]{\textcolor{blue}{#1}} 
\usepackage{dirtytalk}

\title{Hyperspherical Forward-Forward with Prototypical Representations}

\author{Shalini Sarode$^{1,2}$\thanks{Equal contribution.} \quad Brian Moser$^{1}$$^\star$ \quad Joachim Folz$^{1}$ \quad Federico Raue$^{1}$ \\
\textbf{Tobias Nauen}$^{1,2}$ \quad \textbf{Stanislav Frolov}$^{1}$ \quad \textbf{Andreas Dengel}$^{1,2}$ \\
\small $^1$German Research Center for Artificial Intelligence (DFKI), $^2$RPTU Kaiserslautern-Landau \\
{\tt\small first.last@dfki.de}
}

\begin{document}

\maketitle

\begin{abstract}
  The Forward-Forward (FF) algorithm presents a compelling, bio-inspired alternative to backpropagation. 
However, while efficient in training, it has a computationally prohibitive inference process that requires a separate forward pass for every class that is evaluated. 
In this work, we introduce the Hyperspherical Forward-Forward (HFF), a novel reformulation that resolves this critical bottleneck. 
Our core innovation is to reframe the local objective of each layer from a binary goodness-of-fit task to a direct multi-class classification problem within a hyperspherical feature space. We achieve this by learning a set of class-specific, unit-norm prototypes that act as geometric anchors and implicit negatives. 
This architectural innovation preserves the benefits of local training while enabling weight update and inference in a single forward pass, making it \textgreater40x faster than the original FF algorithm. 
Our method is simple to implement, scales effectively to modern convolutional architectures, and achieves superior accuracy on standard image classification benchmarks, closing the gap with backpropagation. Most notably, we are among the first greedy local-learning methods to report over 25\% top-1 accuracy on ImageNet-1k, and 65.96\% with transfer learning.
\end{abstract}

\section{Introduction}
The remarkable success of deep neural networks is fundamentally tied to backpropagation~\cite{rumelhart1986learning,kinga2015method,loshchilov2017decoupled}. 
For decades, it has been the de facto standard for training and enabled models to learn complex hierarchical representations from vast amounts of data~\cite{krizhevsky2012imagenet,lecun2002gradient,caron2021emerging, kaplan2020scaling, nakkiran2021deep}. 
Yet, this reliance on end-to-end backpropagation is not without costly consequences \cite{kaplan2020scaling, bengio1994learning, hochreiter1997long, adam2014method}. 
Its sequential nature, which requires a full forward pass followed by a full backward pass, introduces substantial computational latency and memory requirements \cite{scellier2017equilibrium, hu2022lora, zhang2023adding, kudiabor2024ai, wan2024bridging}. 
These costs increase as models scale in depth, complexity, and size~\cite{he2016deep,vaswani2017attention, nakkiran2021deep, oquab2023dinov2, assran2023self}. 
For what it is worth, the need to transmit precise, real-valued gradients backward through the entire network presents a stark divergence from the local, asynchronous learning mechanisms observed in biological neural systems~\cite{crick1989recent, friston2010free, graves2004biologically, hebb2005organization, xiao2018biologically}. 

To address these challenges, recent work has explored alternative training paradigms \cite{hinton2022forward, krutsylo2025scalable, lee2023symba, journe2022hebbian, dooms2023trifecta}. 
The most promising among these is the Forward-Forward (FF) algorithm~\cite{hinton2022forward}, which replaces the backward pass with a second forward pass. 
In FF, each layer is trained greedily on a local objective: to distinguish ``positive'' data (\emph{i.e.}, images with correct labels) from ``negative'' data (\emph{i.e.}, images with incorrect labels). 
This local, layer-wise training elegantly sidesteps the burdens of backpropagation. 

However, the original FF algorithm suffers from an inefficient inference procedure \cite{bartunov2018assessing, aminifar2024lightff, krutsylo2025scalable,gong2025mono}, a critical limitation that has hindered its practical adoption.
To classify a given test image, the network must perform a separate, full forward pass for every possible class, calculating a ``goodness'' score for each and subsequently returning the class with the highest score. 
This classification process scales linearly with the number of classes, rendering inference computationally prohibitive for datasets with many classes such as CIFAR-100 \cite{CIFAR100} or ImageNet-1K \cite{deng2009imagenet}. 

\begin{figure*}[t]
    \begin{minipage}[b]{0.50\linewidth}
    \centering
  \includegraphics[width=\linewidth]{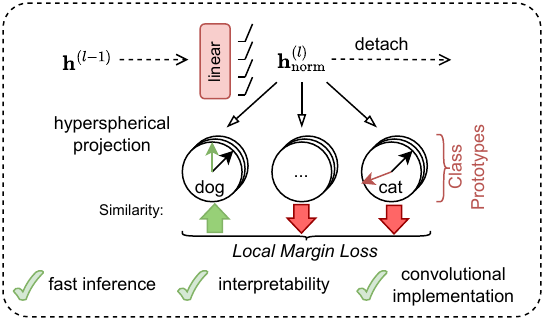}
  \end{minipage}
  \hfill
  \begin{minipage}[b]{0.48\textwidth}
  \caption{\textbf{Visualization of Hyperspherical Forward–Forward (HFF).}
  Each layer produces an activity vector $\mathbf{h}^l$ and applies $\ell_2$ normalization to project features onto the unit hypersphere, compares them to a small set of class prototypes (colored vectors) via a similarity measure, aggregates with LogSumExp, and optimizes a strictly local loss. Layers are trained greedily with their predecessors frozen. At test time, logits are available without per-class passes, enabling \emph{single-pass} inference. Our hyperspherical formulation extends naturally to convolutional blocks.}
  \label{fig:fig1_teaser}
  \end{minipage}
\end{figure*}

In this paper, we introduce Hyperspherical Forward-Forward (HFF), a novel reformulation of the FF algorithm that resolves this critical inference bottleneck. 
HFF reframes the local learning objective of each layer from a binary goodness discrimination task to a direct, multi-class classification task.
It achieves this by projecting layer outputs onto a unit hypersphere and learning a set of class-specific prototypes that serve as geometric anchors in feature space. 
This architectural shift inspired by proxy learning\cite{kim2020proxy}, preserves the desirable properties of local, layer-wise training while enabling highly efficient, single-pass inference. 
For any given input, our model produces a vector of goodness scores for all classes at each layer, allowing the final prediction to be determined in a single forward pass. 
This makes HFF orders of magnitude faster than the original FF algorithm and thus extensible to modern convolutional architectures. 
In our experiments, we show that HFF exceeds the performance of standard FF by 16 percentage points (p.p) on CIFAR10 while reducing inference time by a factor equivalent to the number of classes.

\section{Related Works}
 Backpropagation has long been the backbone of deep learning, yet it is widely regarded as biologically implausible \cite{song2020can, whittington2019theories, lillicrap2020backpropagation, hinton2022forward, journe2022hebbian}. 
It requires a full computational graph and exact weight symmetry for gradient computation, which are assumptions that natural neural systems likely do not satisfy.
This gap between artificial and biological learning has inspired a range of alternatives discussed in the following.

\textbf{Biologically-Inspired Learning.} 
Efforts to improve biological plausibility often target the symmetry requirement in feedback. 
Predictive coding \cite{millidge2022predictive} continuously predicts sensory input and adjusts representations through iterative inference, but this process is computationally costly. 
Feedback alignment (FA) \cite{lillicrap2014random} removes weight symmetry by using fixed random feedback connections, and direct feedback alignment (DFA) \cite{nokland2016direct} further simplifies these updates. 
SoftHebb \cite{journe2022hebbian} discards feedback entirely, relying instead on local Hebbian rules. 
Despite these advances, such models still depend on forms of a backward pass.

\textbf{Forward-Forward (FF).}
The FF algorithm \cite{hinton2022forward} replaces backpropagation with two forward passes and local objectives. 
Each layer maximizes a “goodness” function for positive samples and minimizes it for negative ones. 
While conceptually elegant, FF faces key challenges: (1) explicit negative sample generation is ad hoc and dataset-specific; (2) the loss relies on a discrete threshold rather than a smooth objective; and (3) inference scales linearly with the number of classes, making large-scale tasks impractical.

Subsequent works have tried to address these issues. 
SymBa \cite{lee2023symba} introduces a symmetric softplus-based loss, removing the threshold and encoding labels as intrinsic binary patterns rather than one-hot vectors. 
Scodellaro \textit{et al.}~\cite{scodellaro2023training,scodellaro2025training} extend FF to convolutional layers via spatial label maps, though its generalization remains limited. 
Other variants, such as FFCL~\cite{karkehabadi2024ffcl} and Layer Collaboration \cite{lorberbom2024layer}, incorporate inter-layer communication.  Dendritic Localized Learning (DLL) \cite{lv2025dendritic} advances biological realism with asymmetric feedback and purely local updates. 
Trifecta FF \cite{dooms2023trifecta} jointly contrasts multiple positive and negative samples, improving robustness but increasing computation. 
No-Prop \cite{li2025noprop} removes negative sampling entirely, learning via forward modulation signals, but struggles with deep architectures. 
CwConv \cite{papachristodoulou2024convolutional} improves small-scale performance but lacks scalability due to class-wise partitioning of features requiring wider networks as the number of classes increases. 
Scalable FF (SFF)~\cite{krutsylo2025scalable} explores a novel margin-based local loss and employs auxiliary $1{\times}1$ convolutions (AuxConvs) and pooling to create class-wise goodness tensors. 
Its reliance on AuxConvs to classify confines SFF to CNNs. MonoForward (MF)\cite{gong2025mono}, similar to the backpropagation-free version of Local Errors\cite{nokland2019training}, directly projects layer activities onto the logit space and computes the local loss. SFF and MF have not been explored on large benchmarks such as ImageNet-1K~\cite{deng2009imagenet}. 
\textbf{Our Contribution.}  
In this work, we introduce Hyperspherical Forward-Forward (HFF), which reframes the local error as a direct multi-class classification task in a hyperspherical feature space. Each layer learns a set of class-specific, unit-norm prototypes that serve as geometric anchors for the feature vector. HFF applies uniformly to Multi-Layer Perceptrons (MLPs) and CNNs. It also scales to large, complex datasets with many classes and has per-layer predictions offering greater interpretability.

\section{Methodology}
In this section, we provide a brief formal overview of the FF \cite{hinton2022forward} algorithm and introduce our Hyperspherical Forward-Forward (HFF), including its objective, training phase, and inference.
Further information, such as gradients of the local objectives, an information-theoretic analysis, or extension to convolutional operations can be found in the appendix.
\subsection{Preliminaries: Forward-Forward (FF)}
\label{sec:ff_preliminaries}
FF~\cite{hinton2022forward} optimizes a neural network in a layer-wise greedy manner. 
Each layer $l$ is optimized on a local objective designed to distinguish between ``positive'' and ``negative'' data. 
A positive sample $\mathbf{x}^{\text{pos}}$ consists of a real input (\emph{e.g.}, an image) paired with its true label, while a negative sample $\mathbf{x}^{\text{neg}}$ pairs the same input with an incorrect label.

For each layer, the goal is to learn a mapping that produces a high ``goodness'' score for positive data and a low score for negative data. 
The goodness of a layer's output vector $\mathbf{h}^{(l)}$ is typically defined as the squared norm of the activity vector:
\begin{equation}
    g(\mathbf{h}^{(l)}) = \sum_{i} (h_i^{(l)})^2 = || h^{(l)} ||^2_2.
\end{equation}

The parameters of the layer are then updated to maximize the goodness for positive inputs $\mathbf{h}_{\text{pos}}^{(l)}$ and minimize it for negative inputs $\mathbf{h}_{\text{neg}}^{(l)}$. 
This is achieved by minimizing a local loss function, formulated as a binary classification task using a logistic loss against a fixed threshold $\theta$:
\begin{equation}
\label{eq:FFLoss}
    \mathcal{L}_{\text{FF}}^{(l)} = \log(1 + e^{- (g(\mathbf{h}_{\text{pos}}^{(l)}) - \theta)}) + \log(1 + e^{+ (g(\mathbf{h}_{\text{neg}}^{(l)}) - \theta)}).
\end{equation}

Once a layer is trained, its weights are frozen and its activations are passed to the next layer.  
To prevent later layers from trivially exploiting vector magnitude, FF normalizes the activations before forwarding them.  
This normalization step removes scale information so that only the \textit{direction} of activations is propagated, forcing each subsequent layer to learn meaningful new representations rather than rely on inherited signal magnitude. 

\subsection{Hyperspherical Forward-Forward (HFF)}
Our method adheres to the layer-wise training principle of the FF algorithm but reformulates the local objective from a binary discrimination task to a direct multi-class classification. 
This design shift preserves the benefits of local training while enabling a highly efficient, single-pass inference procedure. 
Moreover, it eliminates the primary inference bottleneck of the original FF algorithm and allows for a straightforward extension to CNNs.

\textbf{Hyperspherical Projection}
For each layer $l$, the forward pass begins with a standard affine transformation of its local input activity vector $\mathbf{h}^{(l-1)}$ (with $\mathbf{h}^{(0)}=\mathbf{x}$) using a weight matrix $\mathbf{W}^{(l)}$ and a bias vector $\mathbf{b}^{(l)}$:
\begin{equation}
\label{eq:affine}
    \mathbf{f}^{(l)} = \mathbf{W}^{(l)}\mathbf{h}^{(l-1)} + \mathbf{b}^{(l)}.
\end{equation}
The local output activity vector is then given by the non-linear projection of \autoref{eq:affine}:
\begin{equation}
\label{eq:decoup}
    \mathbf{h}^{(l)} = \sigma(\mathbf{f}^{(l)})
\end{equation}
where $\sigma$ is a non-linear activation function like \texttt{ReLU} \cite{agarap2018deep}.
We then normalize the output vector $\mathbf{h}^{(l)}$ to have a Euclidean norm of 1 by projecting the vector onto the surface of a unit hypersphere $\mathcal{S}^{D-1}$, where $D$ is the dimensionality of the layer's output space:
\begin{equation}
\label{eq:affineNorm}
    \mathbf{h}_{\text{norm}}^{(l)} = \frac{\mathbf{h}^{(l)}}{\|\mathbf{h}^{(l)}\|_2}.
\end{equation}

\textbf{Class-Specific Prototypical Learning}

Instead of learning a contrastive goodness score, HFF trains the model to classify the input at every layer. 
To achieve this, each layer $l$ maintains a dedicated set of $P$ learnable, unit-norm \textit{prototypes} for each of the $C$ classes, denoted $\{\mathbf{v}_{c,p}^{(l)}\}_{p=1}^P$ for each class $c \in \{1, \dots, C\}$. 
These class-specific prototypes act as anchors in the hyperspherical feature space, defining a unique geometric region for each class \cite{kim2020proxy, ma2024interpretable, nauta2023pip, hou2022closer, gupta2023class}.

From the normalized output $\mathbf{h}_{\text{norm}}^{(l)}$, we compute a vector of class scores $\mathbf{g}_{\text{scores}}^{(l)} \in \mathbb{R}^C$. 
Each component $c$ of this vector represents the goodness for that class, calculated as the soft maximum of the \textit{similarities} between $\mathbf{h}_{\text{norm}}^{(l)}$ and all $P$ prototypes belonging to class $c$. 
We use the LogSumExp (LSE) \cite{gladin2025improved, yao2025logsumexp, miyagawa2021power} function for this purpose:
\begin{equation}
\label{eq:LSE}
    \mathbf{g}_{\text{scores}}^{(l)}[c] = \frac{1}{\tau} \log \sum_{p=1}^{P} e^{\tau \cdot ((\mathbf{h}_{\text{norm}}^{(l)})^T \mathbf{v}_{c,p}^{(l)})}.
\end{equation}
The temperature $\tau$ controls how ``max-like'' the LSE is,  \emph{i.e.} when $P>1$, $\tau$ interpolates between the nearest-prototype matching (high $\tau$) and the class-average similarity  (low $\tau$). When $P=1$, $\tau$ has no effect.
This formulation produces a vector of logits for each input, which directly feeds into a local classification loss.

\textbf{Local Loss} 
HFF trains each layer using its own local objective. 
This involves maximizing the positive score (similarity with the true prototype) \(\mathbf{g}_{\text{scores}}^{(l)}[\mathbf{y}]\), where \(\mathbf{y}\) represents the true-label, and minimizing the scores of the remaining negative classes \(\mathbf{g}_{\text{scores}}^{(l)}[\neg \mathbf{y}]\). To achieve this, we use a contrastive smooth-margin loss (\emph{i.e.}, without a threshold) as described in \cite{krutsylo2025scalable,lee2023symba}: 
\begin{equation}
    \label{eq:HFF_smooth_margin_loss}
    \mathcal{L}_{\text{HFF}}^{(l)} = \log(1 + e^{-(\mathbf{g}_{\text{scores}}^{(l)}[\mathbf{y}] - \text{LSE}(\tau)(\mathbf{g}_{\text{scores}}^{(l)}[\neg \mathbf{y}])})
\end{equation} 
This loss \textit{pulls} the activity vector closer to the true-class prototype and \textit{pushes} it away from all the negative anchors. Here, a larger $\tau$ makes LSE behave like a hard max (focusing on the closest negative anchor), while a smaller $\tau$ averages all negatives.

\textbf{Training}
The parameters $\{\mathbf{W}^{(l)}, \mathbf{b}^{(l)}, \{\mathbf{v}_{c,p}^{(l)}\}_{p=1}^P\}_{l=1}^L$ are optimized in a greedy, sequential manner. 
During the training step of layer $l$, we freeze the parameters of all preceding layers ($k<l$). 
The input $\mathbf{h}^{(l-1)}$ to layer $l$ is first forward-propagated. 
Then, we compute the local loss $L_{\text{HFF}}^{(l)}$ and update only the parameters of layer $l$. 
Critically, we \textit{detach} the activated output $\mathbf{h}^{(l)}$ from the computational graph before passing it to the next layer, thereby ensuring strictly local optimization.

\textbf{Inference}
Our approach makes inference a highly efficient, \textit{single forward pass}. 
For a given test image $\mathbf{x}$, we propagate it through the network once. 
We obtain a final score vector by selecting the scores from the last layer:
\begin{equation}
    \mathbf{G}_{\text{last}}(\mathbf{x}) = \mathbf{g}_{\text{scores}}^{(L)}.
\end{equation}
This process is substantially faster than the original FF formulation, which requires a separate, full forward pass for every candidate class.
The predicted label $\hat{c}$ is simply the class corresponding to the maximum value. 
More formally,
\begin{equation}
    \hat{c} = \underset{c}{\arg\max} \, \mathbf{G}_{\text{last}}(\mathbf{x})[c].
\end{equation}
Additionally, at each layer $l$, we can compute argmax the vector of class scores $\mathbf{g}_{\text{scores}}^{(l)}$ and obtain layerwise predictions. 

\textbf{Comparison regarding Convolution}
Our approach offers a significant advantage in simplicity over conventional FF. 
In more detail, the original formulation requires labels to be embedded within the input. 
These can include adding extra channels for labels alone or designing complex, spatially-extended label masks that interfere with the natural operation of convolution \cite{scodellaro2025training, dooms2023trifecta, wu2024distance, torres2025advancements}. In contrast, our method completely decouples the local objective from the feature extraction. 
Our HFF convolution operates on the image alone, producing a feature map that is then used to derive the loss. 
This separation allows for a much more straightforward application of local, backpropagation-free learning to standard CNNs and other architectures in general, eliminating the need for ad-hoc label engineering.

\textbf{On Intra-Layer Backpropagation}
While our proposed method circumvents end-to-end backpropagation \textit{between} layers, it is worth mentioning that a local gradient computation still occurs \textit{within} each layer. 
To update the weights of a layer $l$ ($\{\mathbf{W}_{\text{conv}}^{(l)}$, $\mathbf{b}_{\text{conv}}^{(l)}\}$ or $\{\mathbf{W}^{(l)}$, $\mathbf{b}^{(l)}\}$) and its prototypes ($\mathbf{v}_{c,p}^{(l)}$), the gradients flow from the local loss function in \autoref{eq:HFF_smooth_margin_loss} back through the LSE and hyperspherical similarity calculations in \autoref{eq:LSE} and finally reaches \autoref{eq:affineNorm} and \autoref{eq:affine}. 

\section{Experiments}
In this section, we test our approach with state-of-the-art FF algorithms and analyze our proposed HFF method regarding different design choices and runtime, as well as transfer learning and explainability.\footnote{Code will be released after the review period.}

\begin{table*}[ht]
\centering
\vspace{-0.5pt}
\caption{Accuracy (\%) of HFF-MLP compared to state-of-the-art local learning methods.
\textbf{Bold} marks the best, \underline{underlined} the second-best among local-loss methods, and \hb{blue} the overall best (except backpropagation). \cmark methods rely on explicit negatives, which are ad hoc and require multiple passes for inference, while methods with a \xmark have implicit negatives and single-pass inference.}

\label{tab:main_result_mlp}
\resizebox{0.8\linewidth}{!}{
\begin{tabular}{lclll}
\toprule
\textbf{Method} & \textbf{Neg.} & \textbf{MNIST} & \textbf{fMNIST} & \textbf{C10} \\
\midrule
\multicolumn{5}{l}{\textit{Backprop baseline}} \\[1pt]
MLP & \xmark & 98.69 \small±0.04 & 89.63 \small±0.25 & 62.53 \small ±0.25 \\[2pt]

\midrule 
\multicolumn{5}{l}{\textit{Hebbian and asymmetric feedback}} \\[2pt]
Hebb Learning~\cite{hebb2005organization}        & \xmark & 78.29 & 67.40 & 19.98 \\
FA~\cite{bartunov2018assessing} & \xmark & 91.87 & 82.16 & 48.46 \\[2pt]
\midrule
\multicolumn{5}{l}{\textit{Forward-only}} \\[1pt]
FF~\cite{hinton2022forward}                 & \cmark & 96.99 & --     & 50.27 \\
FFCL~\cite{karkehabadi2024ffcl}                 & \cmark & 97.23 & 89.71 & 49.93 \\
Layer Collab~\cite{lorberbom2024layer}          & \cmark & 97.90 & 88.40  & 48.40 \\
SymBa~\cite{lee2023symba}                       & \cmark & 98.58 & --   & \underline{59.09} \\
FFCM~\cite{aghagolzadeh2024marginal}            & \cmark & 97.12 & 87.64  & 54.48 \\
\cmidrule(lr){1-5}
Equilibrium Prop~\cite{scellier2017equilibrium}     & \xmark & 93.81 & 75.65 & 16.93 \\
DLL~\cite{lv2025dendritic}   & \xmark & 97.57 & 87.50 & 45.87 \\
MF~\cite{gong2025mono} & \xmark & \hb{\textbf{98.74}} & \hb{\textbf{90.51}} & 56.40 \\

\rowcolor{gray!20}
\textbf{HFF-MLP (ours)}     & \xmark & \underline{98.59} \small±0.07 & \underline{89.96} \small±0.10 & \hb{\textbf{61.93}}  \small±0.23\\
\bottomrule
\end{tabular}
}
\end{table*}

\begin{table*}[t]
\centering
\caption{Accuracy (\%) of CNN-based local learning methods. 
\textbf{Bold} marks the best and \underline{underlined} the second-best among local-loss methods per dataset; 
\hb{blue} highlights the overall best (except BP). ${\dagger}$:intra-layer BP}
\label{tab:main_result_cnn}
\resizebox{0.9\linewidth}{!}{
\begin{tabular}{llllc}
\toprule
\textbf{Method} & \textbf{MNIST} & \textbf{C10} & \textbf{C100} & \textbf{IN1k} \\
\midrule
\multicolumn{5}{l}{\textit{Backprop baseline}} \\
CNN & 99.25 \small±0.08 & 84.37 \small±0.34 & 55.16 \small±0.25 &  71.59 \\[1pt]

\midrule
\multicolumn{5}{l}{\textit{Hebbian and asymmetric feedback}} \\[1pt]
SoftHebb (unsup)~\cite{journe2022hebbian} & - & 80.30 & \hb{56.00} & \hb{27.83} \\
Hebb Learning~\cite{hebb2005organization} & 83.05 & 29.86 & - & - \\
FA~\cite{bartunov2018assessing} & 97.00 & 72.90 & 39.50 & 6.90 \\
DFA~\cite{nokland2016direct} & 98.90 & 72.30 & 43.70 & 6.20 \\
SSDFA~\cite{crafton2019direct} & 98.80 & 73.10 & 41.80 & 2.80 \\[3pt]

\midrule
\multicolumn{5}{l}{\textit{Forward-only}} \\[1pt]
CNN FF (neg)~\cite{scodellaro2023training,scodellaro2025training} & 98.74 & 68.60 & 38.20 & - \\
DLL~\cite{lv2025dendritic} & 98.87 & 70.89 & 38.60 & - \\
CwConv~\cite{papachristodoulou2024convolutional} & 99.42 & 78.11 & 51.23 & - \\
Eq. Prop~\cite{scellier2017equilibrium} & 26.73 & 10.32 & - & - \\
No-Prop-DT~\cite{li2025noprop} & 99.42 & 78.11 & 46.06 & - \\
MF~\cite{gong2025mono} & \hb{\textbf{99.74}} & \underline{82.39} & \textbf{54.77} & - \\
$\text{TFF}^{\dagger}$~\cite{dooms2023trifecta} & \underline{99.53} & 80.01 & 35.79 & - \\
$\text{SFF}^{\dagger}$~\cite{krutsylo2025scalable}  & - & 70.34 & 45.79 & - \\
\rowcolor{gray!20}
$\textbf{HFF-CNN (ours)}^{\dagger}$  & 99.09 \small±0.05 & \hb{\textbf{83.08}} \small±0.45 & \underline{54.34} \small±0.50 & \textbf{25.70} \\
\bottomrule
\end{tabular}}
\end{table*}
\subsection{Comparison with State-of-the-Art}
We evaluate our Hyperspherical Forward-Forward (HFF) model against state-of-the-art backpropagation-free methods across five benchmark datasets. Results for HFF-MLP and its CNN variant are summarized in \autoref{tab:main_result_mlp} and \autoref{tab:main_result_cnn}. Each result except ImageNet1k shows the average and standard deviation of 3 independent trials. 
The results show that HFF consistently delivers superior accuracy as dataset complexity increases, outperforming other backpropagation-free methods and demonstrating the clear advantage of its prototype-based, direct multi-class learning objective. 

\textbf{MNIST and fMNIST.} On MNIST \cite{deng2012mnist} and FashionMNIST~\cite{xiao2017fashion}, datasets with low intra-class variance, our HFF model achieves a strong accuracy of 98.59\% and 89.83\% with MLP, surpassing most methods but falling slightly behind MF~\cite{gong2025mono}.
HFF-CNN achieves 99.08\%, a comparable result to backprop and specialized CNN methods \cite{dooms2023trifecta,krutsylo2025scalable,papachristodoulou2024convolutional} that do not have an MLP counterpart.
The primary benefits of our geometric approach become more pronounced on more challenging tasks. 

\textbf{CIFAR.} HFF's advantage becomes clear on CIFAR-10 (C10) and CIFAR-100 (C100)~\cite{CIFAR100}. 
On CIFAR-10, HFF achieves 61.93\% (MLP) and 83.08\% (CNN) accuracy, establishing a new state-of-the-art for local-loss methods and notably outperforming the strong SymBa MLP which requires explicit negatives and MF CNN baselines by 3 and 1 p.p, respectively. 
The performance gap widens significantly on the more complex CIFAR-100 dataset, where HFF reaches 54.34\% accuracy, second only to MF, and just 0.82 p.p behind the CNN-BP baseline.  
This demonstrates that as the number of classes and the visual complexity grow, HFF's ability to learn separable, prototype-anchored representations provides a distinct advantage over methods based on a binary goodness objective. 

\textbf{ImageNet.} Finally, we test the scalability of HFF on the ImageNet-1K (IN1k)~\cite{deng2009imagenet} dataset using a VGG16 architecture. Although SimCNN~\cite{belilovsky2019greedy} achieves 58.1\% accuracy on 112$\times$112 downsampled images with BP in auxiliary classifiers,
to the best of our knowledge, we are among the \textit{first} local-loss methods to report Top-1 accuracy on the full ImageNet-1K dataset, achieving 25.70\%. HFF trails behind SoftHebb by 2 p.p, but is more than 3x the accuracy of other backpropagation-free credit assignment methods like Feedback Alignment~\cite{bartunov2018assessing}, which use an AlexNet~\cite{krizhevsky2012imagenet} architecture. 
While the absolute performance reflects the known difficulty of training large-scale models without end-to-end backpropagation, we can close the gap by using transfer learning~\cite{crafton2019direct}, further described in \autoref{sec:transfer_learning}.
This result underscores that HFF is more accurate and provides a more viable and scalable path forward for training CNNs without relying on a global backward pass.

\subsection{Inference and Convergence Time}
\label{sec:wallclock}
We benchmark inference on CIFAR-100 (validation set) with a batch size of 1 on an NVIDIA RTX~3090, where FF and HFF use a 2k$\times$3 MLP while the BP model uses a 2k$\times$3-100 MLP (see \autoref{tab:cifar100_ff_timing}). 
Baseline~FF\cite{hinton2022forward} performs $C$ class-conditioned forward passes per input, resulting in an $\mathcal{O}(C)$ inference cost, whereas HFF computes class logits in a single pass, \emph{i.e.}, $\mathcal{O}(1)$ in the number of classes.
In practice, the baseline~FF takes approximately $40\times$ longer than HFF and BP under this configuration. The gap between the theoretical ($100\times$) and observed speedup ($40\times$) arises from GPU parallelism. 
We also benchmark the time to convergence during training for all three methods on MNIST dataset. In this experiment, we record the time and number of epochs needed to consistently perform above a set target (\emph{i.e.} reach validation accuracy $\ge$ 95\% for at least 3 consecutive epochs). While HFF requires 1.7$\times$  BP train time, FF takes 9.5$\times$ longer to converge than BP.  

\begin{table}[ht]
\vspace{-1mm}
\centering
\caption{\textbf{1.}  Wall-clock inference time on C100 (val set) with batch size 1 on an NVIDIA RTX 3090 and 1 dataloader worker. \textbf{2.} MNIST train time-to-convergence (Time to reach target validation accuracy $\ge$ 95\%, for at least patience=3-epochs). All results are averaged over 3 trials.}
\label{tab:cifar100_ff_timing}
\small\resizebox{0.65\linewidth}{!}{
\begin{tabular}{c c c c}
\toprule
\textbf{Method} & \textbf{Time (s)} & \textbf{Throughput (img/s)} & \textbf{Latency (ms/img)} \\
Backprop & 12.99 ± 0.99 & 772.15 ± 56.34 & 1.29 ± 0.10 \\
Baseline FF & 	565.47 ± 15.21 & 17.69 ± 0.47 & 56.54± 1.51 \\
\textbf{HFF-MLP} & \textbf{14.39 ± 1.51} & \textbf{699.21 ± 69.58} & \textbf{1.43 ± 0.15} \\
\midrule
\textbf{Method} & \textbf{Time (s)} & \textbf{Epochs} & \textbf{Best Acc} \\
BackProp & 19.86 ± 3.76 & 3.33 ± 0.58 & 97.37 ± 0.45 \\
Baseline FF  & 188.76 ± 9.21 & 26.67 ± 1.15 & 95.25 ± 0.20 \\
\textbf{HFF-MLP} & 34.23 ± 0.87 & 4.00 ± 0.00 & 96.57 ± 0.29 \\[1pt]
\bottomrule
\end{tabular}}
\end{table}\vspace{-1mm}

\subsection{Ablation Study}
We conduct a series of ablation experiments to evaluate individual components within our framework. 
The configuration corresponding to a \colorbox{gray!20}{gray} cell is our design choice for subsequent sections.

\textbf{Post- and Pre-Activation}
HFF allows flexibility in where the local loss is applied within each layer. 
In the standard setup, we compute it on the \textit{post-activation} features $\mathbf{h}_{\text{norm}}^{(l)}$ (as defined in \autoref{eq:decoup}), \emph{i.e.}, after applying nonlinearity. 
Alternatively, we can calculate the loss on the \textit{pre-activation} features $\mathbf{f}_{\text{norm}}^{(l)}$ and apply the nonlinearity just before the next layer, keeping the gradient flow within the linear block.
This ablation examines how this choice affects performance and the feasibility of bounded and non-differentiable activations, such as \texttt{tanh} or \texttt{step}.
Pre-activation training is faster, enables such activations, and performs better than FF. Still, it consistently underperforms the standard post-activation formulation, indicating that nonlinearity yields more stable learning dynamics. Although \texttt{abs} performs better on MNIST, \texttt{ReLU} is more consistent and better on CIFAR-10.
\autoref{tab:ablation_activation} summarizes the results.

\begin{table*}
\begin{minipage}[b][0.4\textheight]{0.58\linewidth}
\begin{minipage}[\textheight]{\linewidth}
\caption{Ablation on loss placement relative to activation.
\textit{Post-activation} (standard) computes the loss on $\mathbf{h}_{\text{norm}}^{(l)}$ after the nonlinearity, 
while \textit{pre-activation} applies it directly to $\mathbf{f}^{(l)}$ before activation. 
Test accuracy (\%) across datasets for different activation functions.}
\label{tab:ablation_activation}
\vspace{8mm}
\resizebox{\linewidth}{!}{%
\begin{tabular}{cccccccc}
\toprule
 & \textbf{Dataset} & \textbf{Activation} &  \texttt{\textbf{ReLU}} & \texttt{\textbf{Tanh}} & \texttt{\textbf{Sin}} & \texttt{\textbf{Abs}} & \texttt{\textbf{Step}} \\
\midrule
\multirow{4}{*}{MLP} & \multirow{2}{*}{MNIST} & Post & \cellcolor{gray!20}98.53  &  98.64 & \textbf{98.70}  & 98.59 & - \\
& & Pre &  97.25 & 96.66  & 96.48 & \textbf{97.4} &   95.44  \\
\cmidrule(lr){2-8}
& \multirow{2}{*}{CIFAR-10} & Post & \cellcolor{gray!20}\textbf{61.33} & 44.36 & 46.93  & 52.85 & - \\
& & Pre & \textbf{46.01} & 37.08 & 36.57 &  32.08 & 37.07 \\
\midrule
\multirow{4}{*}{CNN} & \multirow{2}{*}{MNIST} & Post & \cellcolor{gray!20}\textbf{98.97}  & 97.86 & 97.96 & 98.76 & - \\
& & Pre & 97.96  &  97.70 & 97.72 & \textbf{98.26} &  95.74   \\
\cmidrule(lr){2-8}
& \multirow{2}{*}{CIFAR-10} & Post &  \cellcolor{gray!20}\textbf{83.18} & 71.51  & 73.47  & 80.65 & - \\
& & Pre &  \textbf{66.92} & 50.10 & 65.11 & 61.09 & 53.53 \\

\bottomrule
\end{tabular}%
}
\end{minipage}%
\vspace{5mm}
\vfill
\vspace{-5mm}
\begin{minipage}[0.2\textheight]{\linewidth}
\caption{
Ablation on the choice of loss function in HFF across architectures and datasets. 
The smooth margin loss consistently outperforms cross-entropy, particularly in deeper CNNs. 
}
\label{tab:ablation_loss}
\resizebox{\linewidth}{!}{%
\begin{tabular}{lccc}
\toprule
\textbf{Loss} & \textbf{MNIST} & \textbf{CIFAR-10} & \textbf{CIFAR-100} \\
\midrule
\multicolumn{4}{l}{\textit{3-Layer MLP}} \\[2pt]
Cross-Entropy & 98.51 & 61.19 & – \\
Smooth Margin & \cellcolor{gray!20}\textbf{98.53} & \cellcolor{gray!20}\textbf{61.33} & – \\[4pt]
\midrule
\multicolumn{4}{l}{\hspace{0.8em}2-Layer CNN (MNIST) \quad / \quad 3-Layer CNN (CIFAR-10/100)} \\[2pt]
Cross-Entropy & 98.82 & 82.13 & 47.67 \\
Smooth Margin & \cellcolor{gray!20}\textbf{98.97} & \cellcolor{gray!20}\textbf{83.18} & \cellcolor{gray!20}\textbf{53.24} \\[4pt]
\midrule
\multicolumn{4}{l}{\textit{11-Layer CNN (VGG11)}} \\[2pt]
Cross-Entropy & – & 73.44 & 49.37 \\
Smooth Margin & – & \cellcolor{gray!20}\textbf{83.48} & \cellcolor{gray!20}\textbf{54.66} \\
\bottomrule
\end{tabular}%
}
\end{minipage}%
\end{minipage}%
\hfill
\begin{minipage}[b][0.3\textheight]{0.39\linewidth}
\begin{minipage}[0.5\textheight]{\linewidth}
\caption{
 Scalability of HFF-MLP on MNIST with respect to network width and depth. 
Increasing either the number of neurons per layer or the number of layers improves accuracy.
}
\label{tab:ablation_scalability}
\resizebox{\linewidth}{!}{%
\begin{tabular}{lc}
\toprule
\textbf{Configuration} & \textbf{Accuracy (\%)} \\
\midrule
\multicolumn{2}{l}{\textit{Width (neurons per layer)}} \\
\textbf{20} & 91.90 \\
\textbf{200} & 94.46 \\
\textbf{500} & 94.54 \\
\textbf{800} & 94.61 \\
\textbf{1100} & \textbf{94.73} \\
\midrule
\multicolumn{2}{l}{\textit{Depth (number of layers)}} \\
\textbf{1024} & 94.65 \\
\textbf{1024-512} & 97.04 \\
\textbf{1024-512-256} & \textbf{97.49} \\
\textbf{2000-2000-2000} & \cellcolor{gray!20}\textbf{98.53} \\
\bottomrule
\end{tabular}%
}
\end{minipage}%
\vfill
\begin{minipage}[0.5\textheight]{\linewidth}
\caption{
Ablation of auxiliary convolution width and pooling scale in HFF-CNN. 
Spatial compression via auxiliary convolutions markedly improves accuracy, with diminishing returns beyond moderate capacity.
}
\label{tab:ablation_auxconv}
\resizebox{\linewidth}{!}{%
\begin{tabular}{ccc}
\toprule
\textbf{Aux} & \multirow{2}{*}{\textbf{CIFAR-10}} & \multirow{2}{*}{\textbf{CIFAR-100}} \\
\textbf{Channels}& &  \\
\midrule
0-0-0 & 73.97 & 38.06 \\
32-16-8 & 82.4 & 46.23 \\
128-64-32 &  \cellcolor{gray!20}\textbf{83.18} & 52.61 \\
256-128-64 &  83.12 & \cellcolor{gray!20}\textbf{53.24} \\
\bottomrule 
\end{tabular}%
 }
 \vspace{-13mm}
 \end{minipage}
\end{minipage}

\end{table*}

\textbf{Loss Function}
The smooth margin loss in \autoref{eq:HFF_smooth_margin_loss} is the standard choice in HFF. 
Unlike the cross-entropy loss, it operates entirely on local logits and directly optimizes the margin between the correct class score and the LSE of all other classes. 
As shown in \autoref{tab:ablation_loss}, the smooth margin consistently outperforms cross-entropy across architectures and datasets, particularly in deeper CNNs, indicating its superior stability and scalability within the HFF training framework.

\textbf{Scalability}
As shown in \autoref{tab:ablation_scalability}, HFF-MLP scales reliably with both model width and depth. 
Increasing the number of neurons per layer yields steady but moderate improvements, indicating that the layer-wise objectives can utilize wider feature spaces effectively even without global gradient flow. 
Increasing depth, however, leads to more pronounced gains, suggesting that the hierarchical composition of locally learned representations remains stable under forward-only training. 
The best result (98.53\%) is achieved with a 3-layer, 2000-neuron-per-layer configuration.

\textbf{Auxiliary Convolutions}
Auxiliary convolutions vary the spatial dimensionality of hidden activations $\mathbf{h}^{(l)}$ and the hypersphere in which we project the vector onto prototypes, unlike the fixed AuxConvs in SFF\cite{krutsylo2025scalable}.
Importantly, these layers are not part of the main feedforward path. 
They only process intermediate activations to form class-wise goodness maps with the prototypes, leaving the network depth unchanged. 
As shown in \autoref{tab:ablation_auxconv}, a 3-layer CNN without auxiliary convolutions (0–0–0) causes a substantial drop in accuracy, indicating their crucial role in providing stable, localized gradients. 
Adding AuxConvs with channels (32–16–8) leads to significant improvement, while moderate channel widths (128–64–32) and (256-128-64) offer better accuracies for CIFAR-10 and CIFAR-100. 
Increasing their capacity further yields minimal or dataset-specific gains, suggesting diminishing returns once sufficient configuration is achieved. 

\begin{table}[t]
\begin{minipage}[b]{0.38\linewidth}
 \caption{Transfer learning from a model trained with backprop to a layer-wise network. The HFF-conv layers are initialized with off-the-shelf ImageNet-1K pretrained weights and frozen, while the last HFF-sphere layers are trainable. Test accuracies(\%) on \textbf{ImageNet-1K} \cite{deng2009imagenet} are compared with other transfer learning methods.}
\label{tab:transfer_learning}
\resizebox{\linewidth}{!}{%
\begin{tabular}{cccc}
\toprule
\textbf{Model} & \textbf{Method} & \textbf{From Scratch} & \textbf{Transfer} \\
\midrule
\multirow{4}{*}{AlexNet} & BP & 56.52 & 49.00 \\
& DFA & 6.20 &  \textbf{48.80} \\
& SSDFA & 2.80 &  46.30 \\
& \cellcolor{gray!20}\textbf{HFF} & \cellcolor{gray!20}\textbf{22.21} & \cellcolor{gray!20}\underline{46.99}  \\
\midrule
\multirow{4}{*}{VGG16}& BP & 71.59 & 65.80 \\
& DFA & -- &  65.30  \\
& SSDFA & -- &  64.50 \\
& \cellcolor{gray!20}\textbf{HFF} & \cellcolor{gray!20}\hb{\textbf{25.70}} & \cellcolor{gray!20}\hb{\textbf{65.96}} \\
\bottomrule
\end{tabular}%
}
\end{minipage}
\hfill
 \begin{minipage}[b]{0.58\linewidth}
 \caption{Ablation of normalizing the activity vector for loss calculation and input to the next layer. (\xmark) denotes normalized vector, $\mathbf{h}_{\text{norm}}^{(l)}$ and (\cmark) denotes vector with length information, $\mathbf{h}^{(l)}$. We ablate these normalization settings in passing to next layer as input and calculating vector similarities with the unit prototypes.  } 
\label{tab:ablation_norm}
\resizebox{\linewidth}{!}{%
    \begin{tabular}{cccccccc}
        \toprule

        \textbf{ $\mathbf{h}^{(l)}$} & \textbf{ $\mathbf{h}^{(l)}$}& \multicolumn{3}{c}{\textbf{MLP}} & \multicolumn{3}{c}{\textbf{CNN}}\\
        \cmidrule{3-8}
         \textbf{Input} & \textbf{Sims}  & MNIST & C10 & C100 & MNIST & C10 & C100 \\
        \midrule
        \xmark & \xmark & 97.83 & 56.75 & \underline{27.41} & 98.97  & \underline{83.18} & 53.24\\ 
        \xmark & \cmark & \underline{98.36} & \underline{57.53} & 25.71 &  \underline{99.04}  &    83.05 & \cellcolor{gray!20}\textbf{54.54}\\
        \cmark & \xmark & 97.85 & 57.30 & \cellcolor{gray!20}\textbf{27.64} & 98.80   & 82.78 & 53.80\\
        \cmark & \cmark & \cellcolor{gray!20}\textbf{98.53} & \cellcolor{gray!20}\textbf{61.33} & 21.37 & \cellcolor{gray!20}\textbf{99.08}  & \cellcolor{gray!20}\textbf{83.25} & \underline{54.35}\\
        \bottomrule
    \end{tabular}%
    }
    \vspace{-14.25mm}
    
 \end{minipage}
\end{table}

\textbf{Normalization}
Although normalizing the activity vector was central to naive FF, HFF has the flexibility to explore using $\mathbf{h}^{(l)}$ directly instead of $\mathbf{h}_{\text{norm}}^{(l)}$, both in the loss calculation and to pass it as input to the next layer. As FF~\cite{hinton2022forward} aims to increase the norm of the positive activity vector and decrease that of the negative activity, passing these vectors without normalizing would be trivial to the next layer, rendering it redundant. Therefore, only the orientation is propagated to subsequent layers, forcing them to learn from scratch. In contrast, our loss is based on angular similarities with unit-norm prototypes. Although the output orientation is sufficient in HFF, the vector's norm information does not impair the next layer's learning. Similarly, we can scale the similarities in \autoref{eq:LSE} with any arbitrary coefficient without loss of calculation. For simplicity, if we choose the vector length as the scaling factor, we get the dot-product $\|\mathbf{h}^{(l)}\|_2(\mathbf{h}_{\text{norm}}^{(l)})^T \mathbf{v}_{c,p}^{(l)}$. In \autoref{tab:ablation_norm}, we ablate these settings for all linear layers in both MLP and CNN across datasets. Scaling the input and similarities gave the best results for MNIST and CIFAR-10. For CIFAR-100, scaling either the input or similarity (but not both) produced the best results, due to more classes.  

\textbf{Transfer Learning and interpretability}
\label{sec:transfer_learning}
Transfer learning (TL) provides an efficient way to adapt traditional models for local learning in resource-constrained environments, such as mobile devices. In this setup, convolutional backbone weights are initialized from a model trained with backpropagation (BP) and kept frozen, while only the final linear layers are fine-tuned.
In prior work \cite{crafton2019direct}, alternative feedback algorithms such as SSDFA\cite{crafton2019direct} and DFA\cite{nokland2016direct} struggle to scale to complex datasets when trained from scratch (achieving only 2–6\% accuracy on AlexNet \cite{krizhevsky2012imagenet}), but attain BP-level performance when used in a transfer learning setup.
Following this approach, we initialize our HFF variant with BP-pretrained convolutional weights and train only the terminal HFF-sphere layers. As shown in \autoref{tab:transfer_learning}, HFF-AlexNet is the second-best, while HFF-VGG surpasses its BP baseline(65.96\% vs 65.80\%), demonstrating the effectiveness of transfer learning in aligning biologically inspired optimization methods with conventional architectures. Furthermore, our greedy learning approach offers better interpretability over the otherwise ``black-box" BP. Analyzing layer-wise accuracies, activation maps, and feature-prototype UMAPs provides richer insights into the model's behavior (see appendix).

\section{Limitations \& Future Work}
Our model relies on a fixed number of learnable prototypes $P$ per class. 
Although this provides flexibility, the optimal value is a hyperparameter that currently requires empirical validation (see Appendix). 
An insufficient number could hinder the representation of complex, multi-modal class distributions. 
Automatically determining or dynamically adapting the number of prototypes during training is a promising direction. Although we randomly initialize and learn the prototypes, future work could explore whether dynamic class-to-prototype assignment methods~\cite{saadabadi2024hyperspherical} can improve our model performance by incorporating inter-class semantics.
Our model achieves strong results without end-to-end feedback. Still, it lags behind large-scale BP models. Closing this gap may require newer techniques to bring global context to local learning. 
Our convolutional extension employs global average pooling and AuxConvs. While effective, this mechanism may create an information bottleneck. It would be valuable to explore other pooling strategies to preserve more spatial information. Finally, our local prototypes require D$\times$P$\times$C additional parameters in each layer. Reducing this memory overhead while maintaining performance would be an interesting direction.
We view these limitations not as shortcomings but as a transparent and honest scoping of our work, highlighting a rich set of opportunities for future extension and refinement.

\section{Conclusion}
In this work, we introduce Hyperspherical Forward-Forward (HFF), a method that retains the efficiency of local, layer-wise training while solving the critical inference bottleneck of the original Forward-Forward (FF) algorithm. 
By reformulating the local objective as a multi-class classification task on a hypersphere using learnable prototypes, we enable fast, single-pass inference. 
Our experiments demonstrate that this approach not only achieves a massive speed-up but also leads to significant accuracy improvements over local learning methods and bridges the gap to backpropagation. 
HFF thus provides a practical and scalable path forward for exploring backpropagation-free learning in modern deep learning architectures.

\textbf{Acknowledgments: }
\small
This work was funded by the Federal Ministry of Research,
Technology, and Space, Germany under project Albatross (16IW24002) and by Carl-Zeiss Foundation under the Sustainable Embedded AI project (P2021-02-009).

\newpage
{
    \small
    \bibliographystyle{unsrtnat}
    \bibliography{bib}
}

\newpage

\appendix
\section{Gradients of Local Objectives}
We provide a mathematical analysis to show that the HFF objective is inherently better suited for learning separable features than the original FF objective. We analyze the gradient of each local loss function with respect to the pre-activation output $\mathbf{f}^{(l)}$, as this gradient dictates the update to the layer's weights $\mathbf{W}^{(l)}$.

For the original FF algorithm, the loss $\mathcal{L}_{FF}^{(l)}$ is a function of the goodness $g(\mathbf{h}^{(l)}) = \|\mathbf{h}^{(l)}\|_2^2$, where $\mathbf{h}^{(l)}$ is the activated output. The gradient with respect to the pre-activation $\mathbf{f}^{(l)}$ is (assuming $\mathbf{f}^{(l)}$ is in the positive domain of a ReLU activation, so $\mathbf{h}^{(l)} = \mathbf{f}^{(l)}$):
\begin{equation}
    \nabla_{\mathbf{f}^{(l)}} \mathcal{L}_{FF}^{(l)} = \frac{\partial \mathcal{L}^{(l)}}{\partial g} \frac{\partial g}{\partial \mathbf{f}^{(l)}} = \frac{\partial \mathcal{L}^{(l)}}{\partial g} \cdot 2\mathbf{f}^{(l)}
\end{equation}
The scalar term $\frac{\partial \mathcal{L}^{(l)}}{\partial g}$ is negative for positive samples and positive for negative samples. Critically, the gradient is proportional to the vector $\mathbf{f}^{(l)}$ itself. A gradient update of the form $\Delta \mathbf{W} \propto \mathbf{f}^{(l)}(\mathbf{h}^{(l-1)})^T$ primarily adjusts the \textit{magnitude} of the output vector $\mathbf{f}^{(l)}$, pushing it to be longer for positive samples and shorter for negative ones, but does not directly alter its direction.

For our HFF algorithm, the loss $\mathcal{L}_{HFF}^{(l)}$ is a function of class scores, which depend on the normalized vector $\mathbf{f}_{\text{norm}}^{(l)}$. The gradient with respect to $\mathbf{f}^{(l)}$ is found via the chain rule:
\begin{equation}
    \nabla_{\mathbf{f}^{(l)}} \mathcal{L}_{HFF}^{(l)} = (\nabla_{\mathbf{f}_{\text{norm}}^{(l)}} \mathcal{L}_{HFF}^{(l)})^T \frac{\partial \mathbf{f}_{\text{norm}}^{(l)}}{\partial \mathbf{f}^{(l)}}
\end{equation}
The Jacobian of the normalization function is $\frac{\partial \mathbf{f}_{\text{norm}}}{\partial \mathbf{f}} = \frac{1}{\|\mathbf{f}\|}(\mathbf{I} - \mathbf{f}_{\text{norm}}\mathbf{f}_{\text{norm}}^T)$. This is a projection matrix that projects any vector onto the subspace orthogonal to $\mathbf{f}_{\text{norm}}$. Consequently, the gradient $\nabla_{\mathbf{f}^{(l)}} \mathcal{L}_{HFF}^{(l)}$ is a vector that is, by construction, orthogonal to the current feature vector $\mathbf{f}^{(l)}$. An update in a direction orthogonal to a vector primarily changes its \textit{direction}, not its magnitude.

This analysis formally demonstrates the different inductive biases. The FF gradient adjusts the vector's magnitude along its current direction. In contrast, the HFF gradient adjusts the vector's direction on the hypersphere, pushing it towards the correct class prototypes and away from incorrect ones. This directly optimizes for angular separability, providing a more effective learning signal for creating discriminative features.

\section{Information-Theoretic Analysis}
We can frame the HFF objective through the lens of the Information Bottleneck principle. The goal is to learn a representation, in our case $\mathbf{z} \equiv \mathbf{f}_{\text{norm}}^{(l)}$, that is maximally informative about the class label $\mathbf{y}$. This is achieved by maximizing the mutual information $I(\mathbf{z}; \mathbf{y})$.
\begin{equation}
    I(\mathbf{z}; \mathbf{y}) = H(\mathbf{y}) - H(\mathbf{y}|\mathbf{z})
\end{equation}
Since the marginal entropy of the labels, $H(\mathbf{y})$, is constant for a given dataset, maximizing $I(\mathbf{z}; \mathbf{y})$ is equivalent to minimizing the conditional entropy $H(\mathbf{y}|\mathbf{z})$. The conditional entropy is defined as:
\begin{equation}
    H(\mathbf{y}|\mathbf{z}) = - \mathbb{E}_{p(\mathbf{z})} \left[ \sum_{\mathbf{y}} p(\mathbf{y}|\mathbf{z}) \log p(\mathbf{y}|\mathbf{z}) \right]
\end{equation}
The true posterior $p(\mathbf{y}|\mathbf{z})$ is unknown. Our model provides a parametric approximation, $q_\phi(\mathbf{y}|\mathbf{z})$, where $\phi$ represents the network parameters (weights and prototypes). The cross-entropy loss used in HFF is an empirical estimate of the negative log-likelihood of this approximation. We can relate this loss to the conditional entropy through the Kullback-Leibler (KL) divergence between the true and approximate posteriors:
\begin{equation}
    D_{KL}(p(\mathbf{y}|\mathbf{z}) || q_\phi(\mathbf{y}|\mathbf{z})) = H(\mathbf{y}|\mathbf{z}, q_\phi) - H(\mathbf{y}|\mathbf{z})
\end{equation}
where $H(\mathbf{y}|\mathbf{z}, q_\phi)$ is the cross-entropy. Rearranging gives:
\begin{equation}
    H(\mathbf{y}|\mathbf{z}) = H(\mathbf{y}|\mathbf{z}, q_\phi) - D_{KL}(p || q_\phi)
\end{equation}
Since $D_{KL}(p || q_\phi) \geq 0$, the cross-entropy provides an upper bound on the true conditional entropy. Minimizing the cross-entropy loss is therefore equivalent to minimizing this upper bound, which serves as a proxy for minimizing the true conditional entropy and thus maximizing the mutual information. The hyperspherical projection acts as the "bottleneck," constraining the capacity of the representation $\mathbf{z}$ and forcing the layer to learn a maximally compressed yet informative mapping.

\section{Dynamics of Prototypical Learning} 
We provide a mathematical proofsketch that the HFF learning objective causes the learnable prototypes, $\{\mathbf{v}_{c,p}\}$, to evolve into geometric centers of their class clusters on the unit hypersphere. 
This is analogous to an online k-means clustering algorithm operating in the angular domain. 
We analyze the gradient of the local loss with respect to the prototypes themselves. 

Consider a single training sample $(\mathbf{x}, y)$ where the ground-truth class is $k$. 
The local cross-entropy loss, $\mathcal{L}_{HFF}^{(l)}$, is minimized by maximizing the score of the correct class, $(\mathbf{g}_{\text{scores}}^{(l)})_k$. 
The gradient of the loss with respect to a single prototype $\mathbf{v}_{k,p}^{(l)}$ belonging to the correct class $k$ is given by the chain rule: 
\begin{equation} 
\nabla_{\mathbf{v}_{k,p}} \mathcal{L}_{HFF} = \frac{\partial \mathcal{L}_{HFF}}{\partial (\mathbf{g}_{\text{scores}})_k} \frac{\partial (\mathbf{g}_{\text{scores}})_k}{\partial \mathbf{v}_{k,p}} 
\end{equation} 
The first term is the derivative of the cross-entropy loss with respect to the logit of the correct class, which is $(p_k - 1)$, where $p_k$ is the softmax probability for class $k$. 
This is a negative scalar value. 
The second term is the derivative of our LogSumExp score function from \autoref{eq:LSE}: \begin{align} 
\frac{\partial (\mathbf{g}_{\text{scores}})_k}{\partial \mathbf{v}_{k,p}} &= \frac{\partial}{\partial \mathbf{v}_{k,p}} \left[ \frac{1}{\tau} \log \sum_{j=1}^{P} e^{\tau \cdot ((\mathbf{f}_{\text{norm}})^T \mathbf{v}_{k,j})} \right] \nonumber \\ 
&= \frac{e^{\tau \cdot ((\mathbf{f}_{\text{norm}})^T \mathbf{v}_{k,p})}}{\sum_{j=1}^{P} e^{\tau \cdot ((\mathbf{f}_{\text{norm}})^T \mathbf{v}_{k,j})}} \cdot \mathbf{f}_{\text{norm}} 
\label{eq:softmax_weight} 
\end{align} 
Let us denote the softmax-like weighting term in \autoref{eq:softmax_weight} as $w_p \in [0, 1]$. 
This weight is largest when the feature vector $\mathbf{f}_{\text{norm}}$ has the highest cosine similarity to the specific prototype $\mathbf{v}_{k,p}$. 
Combining these terms, the full gradient for the prototype is: 
\begin{equation} 
\nabla_{\mathbf{v}_{k,p}} \mathcal{L}_{HFF} = (p_k - 1) \cdot w_p \cdot \mathbf{f}_{\text{norm}}. 
\end{equation} 
Naturally, the gradient descent update rule for the prototype is $\mathbf{v}_{k,p} \leftarrow \mathbf{v}_{k,p} - \eta \nabla_{\mathbf{v}_{k,p}} \mathcal{L}_{HFF}$. 
Substituting our result, the update becomes: 
\begin{equation} 
\Delta \mathbf{v}_{k,p} \propto - (p_k - 1) \cdot w_p \cdot \mathbf{f}_{\text{norm}} \propto  w_p \cdot \mathbf{f}_{\text{norm}} .
\end{equation} 
This update rule demonstrates that the prototype $\mathbf{v}_{k,p}$ is pulled in the direction of the feature vector $\mathbf{f}_{\text{norm}}$ from its own class. 
The magnitude of this pull is weighted by $w_p$, meaning the prototype that is already closest to the feature vector is updated most strongly. 
Over the course of training with many samples, this update rule causes the set of prototypes for each class to converge towards the means of the class-conditional feature distributions on the hypersphere, thus learning to represent the geometric centers of the class clusters.

\section{Lipschitz Continuity and Robustness} 
We expect that each layer in our HFF network is Lipschitz continuous, a property that can guarantee robustness to small input perturbations. 
A function $G$ is $K$-Lipschitz continuous if for any two inputs $\mathbf{x}_1, \mathbf{x}_2$, the inequality $\|G(\mathbf{x}_1) - G(\mathbf{x}_2)\| \le K \|\mathbf{x}_1 - \mathbf{x}_2\|$ holds for a constant $K$. 
This means that bounded changes in the input lead to bounded changes in the output, preventing the amplification of noise or adversarial perturbations. 
We show this by demonstrating that each operation within a HFF layer is Lipschitz continuous. 
The layer's function is a composition of these operations, $G(\mathbf{h}) = (g_3 \circ g_2 \circ g_1)(\mathbf{h})$, where: 
\begin{enumerate} 
\item $g_1(\mathbf{h}) = \mathbf{W}\mathbf{h} + \mathbf{b}$ (Affine transformation) 
\item $g_2(\mathbf{f}) = \mathbf{f} / \|\mathbf{f}\|_2$ (Normalization) 
\item $g_3(\mathbf{f}_{\text{norm}})$ are the class scores from \autoref{eq:LSE}. 
\end{enumerate} 
Since the composition of Lipschitz functions is itself Lipschitz, we can analyze each step. 

\textbf{1. Affine Transformation.} 
The function $g_1(\mathbf{h})$ is an affine map. It is well-known to be Lipschitz continuous with a constant equal to the spectral norm (the largest singular value) of the weight matrix, $K_1 = \|\mathbf{W}^{(l)}\|_2$. 
\begin{equation} 
\|g_1(\mathbf{h}_1) - g_1(\mathbf{h}_2)\| = \|\mathbf{W}(\mathbf{h}_1 - \mathbf{h}_2)\| \le \|\mathbf{W}\|_2 \|\mathbf{h}_1 - \mathbf{h}_2\|. 
\end{equation} 

\textbf{2. Normalization.} 
The function $g_2(\mathbf{f})$ projects a vector onto the unit hypersphere. 
This projection is a non-expansive map, meaning it is 1-Lipschitz ($K_2=1$), provided that $\mathbf{f} \neq \mathbf{0}$. 
\begin{equation} 
\|g_2(\mathbf{f}_1) - g_2(\mathbf{f}_2)\| \le 1 \cdot \|\mathbf{f}_1 - \mathbf{f}_2\|. 
\end{equation} 

\textbf{3. Score Calculation.} 
The final stage, $g_3$, computes scores based on cosine similarities and the LogSumExp function. 
The cosine similarity between a variable unit vector $\mathbf{f}_{\text{norm}}$ and a fixed unit prototype $\mathbf{v}$ is 1-Lipschitz. 
The LogSumExp function is also 1-Lipschitz, as its gradient with respect to its inputs is a probability vector (from the softmax) whose $\ell_2$ norm is less than or equal to 1. 
Therefore, the score calculation step is 1-Lipschitz, $K_3=1$. 

\textbf{Composition.} The Lipschitz constant of the composite function $G$ is bounded by the product of the individual constants: $K_G \le K_1 \cdot K_2 \cdot K_3$. 
For a single HFF layer, this gives: 
\begin{equation} 
K_G \le \|\mathbf{W}^{(l)}\|_2 \cdot 1 \cdot 1 = \|\mathbf{W}^{(l)}\|_2. 
\end{equation} 

In summary, this indicates that each HFF layer is Lipschitz continuous, with the constant determined solely by the spectral norm of its weight matrix. 
This inherent stability arises naturally from the architectural choice of projecting features onto a hypersphere, providing a theoretical foundation for the model's generalization and robustness without requiring explicit regularization techniques.

It is important to note that this stability also extends to the feature representation $\mathbf{h}^{(l)} = \sigma(\mathbf{f}_{\text{norm}}^{(l)})$ that is passed to the next layer. 
Since standard activation functions such as ReLU are 1-Lipschitz continuous, the full mapping from a layer's input to its forward-passed output is also a composition of Lipschitz functions. 
Thus, both the local learning signal and the propagated hierarchical representation benefit from the robustness conferred by the hyperspherical projection.

This observation marks a theoretical advantage over the standard FF algorithm. 
The original FF objective is based on the norm of the output, $g(\mathbf{h}) = \|\mathbf{h}\|_2^2$. 
This goodness function is not Lipschitz continuous, as its gradient, $2\mathbf{h}$, is unbounded. 
This can lead to exploding gradients within the local loss if feature magnitudes become large, resulting in unstable training dynamics. 
In contrast, our HFF replaces this potentially unstable magnitude-based objective with a series of Lipschitz-continuous operations, anchored by the non-expansive normalization step. 
Therefore, the stability of HFF is an inherent architectural property.

\section{Extension to Convolutional Layers}
We extend HFF principles to convolutional layers, which operate on 2D feature maps $\mathbf{h}^{(l-1)} \in \mathbb{R}^{D_{\text{in}} \times H \times W}$. 
Let us assume a standard convolutional operation with a kernel $\mathbf{W}_{\text{conv}}^{(l)}$ that produces an output feature map of the form ${\mathbf{f}^{(l)} \in \mathbb{R}^{D_{\text{out}} \times H' \times W'}}$ and $\mathbf{h}^{(l)}$ after activation.
This map, after pooling, serves as the input to the next layer.

To derive a local loss, we apply global average pooling to this output feature map $\mathbf{h}^{(l)}$.
This operation condenses all spatial information into a single feature vector $\mathbf{z}^{(l)} \in \mathbb{R}^{D_{\text{out}}}$ whose dimension is the number of output channels. 
Additionally, to change the dimension of the hypersphere, $D_{\text{out}}$, we can apply  \textit{auxiliary $1\times1$ convolutions} to produce $\mathbf{h'}^{(l)} \in \mathbb{R}^{D_{\text{out}}' \times H' \times W'}$.
This feature vector is subsequently normalized:
\begin{equation}
    \mathbf{z}_{\text{norm}}^{(l)} = \frac{\text{GlobalAvgPool}(\mathbf{h'}^{(l)})}{\|\text{GlobalAvgPool}(\mathbf{h'}^{(l)})\|_2}.
\end{equation}
 
This normalized vector $\mathbf{z}_{\text{norm}}^{(l)}$ serves the same purpose as $\mathbf{h}_{\text{norm}}^{(l)}$ in the fully-connected layers. 
It is compared against the class prototypes $\{\mathbf{v}_{c,p}^{(l)}\}_{p=1}^P$, whose dimension now matches $D_{\text{out}}$ (or $D'_{\text{out}}$), using LSE to compute the final class scores $\mathbf{g}_{\text{scores}}^{(l)}$. 

\section{Implementation}
\subsection{Datasets}
\textbf{MNIST} consists of 70,000 grayscale images of handwritten digits (0–9) at 28×28 resolution. Its low intra-class variance and simplicity make it a standard benchmark for evaluating fundamental learning dynamics.

\textbf{Fashion-MNIST (FMNIST)} contains 70,000 grayscale images of clothing items across 10 classes, offering a more challenging alternative to MNIST with higher visual complexity while preserving the same format and scale.

\textbf{CIFAR-10} comprises 60,000 natural images across 10 classes at 32×32 resolution. Despite its small size, it features substantial intra-class variation, making it a widely used benchmark for evaluating generalization and robustness.

\textbf{CIFAR-100} expands this challenge to 100 fine-grained categories, greatly increasing task complexity while retaining the same data volume and resolution as CIFAR-10. Its fine-grained structure makes it useful for assessing representation quality.

\textbf{ImageNet-1k} contains 1.28M training and 50,000 validation images spanning 1,000 object categories with high visual diversity and varied image resolutions. It remains the most widely used large-scale benchmark for evaluating the scalability and real-world performance of image classification models.

\subsection{Model Architectures and Hyperparameters}
The models used in our experiments are summarized in \autoref{tab:hff_mlp_configs}, \autoref{tab:hff_cnn_configs} and \autoref{tab:model_archs}. In all our experiments, we use the following settings, unless stated otherwise: standard data augmentation (crop/flip/normalization), a constant LR schedule, Adam Optimizer, EMA updates for the prototypes with a decay of 0.99, and $\tau{=}10$ in the loss function, no BN in convolutional models. 

\begin{table}[ht]
\centering
\caption{HFF-MLP configurations for MNIST, FMNIST, and CIFAR-10.}
\label{tab:hff_mlp_configs}
\resizebox{0.8\linewidth}{!}{
\begin{tabular}{lrrr}
\toprule
\textbf{Config} &
\textbf{MNIST} &
\textbf{FMNIST} &
\textbf{CIFAR-10} \\
\midrule
Learning Rate (lr)          & 0.001 & 0.001 & 0.001 \\
Network Architecture        & 784-2000-2000-2000 & 784-2000-2000-2000 & 784-2000-2000-2000 \\
Number of epochs            & 150 & 150  & 300 \\
Scaled Similarities         & \cmark & \cmark & \cmark \\
Scaled Input                & \cmark & \cmark & \cmark \\
Number of Prototypes        & 1 & 1 & 1 \\
Loss Type                   & Smooth margin & Smooth margin & Smooth margin \\
Activation Type             & ReLU & ReLU & ReLU \\
\bottomrule
\end{tabular}}
\end{table}

\begin{table}[ht]
\centering
\caption{HFF-CNN configurations for MNIST, CIFAR-10,  CIFAR-100.}
\label{tab:hff_cnn_configs}
\resizebox{0.8\linewidth}{!}{
\begin{tabular}{lrrr}
\toprule
\textbf{Config} &
\textbf{MNIST} &
\textbf{FMNIST} &
\textbf{CIFAR} \\
\midrule
Learning Rate (lr)          & 0.0001 & 0.0001 & 0.0001 \\
Network channels            &  1-32-64 & 3-32-64-128 & 3-32-64-128 \\
Auxiliary channels            & 64-32 & 128-64-32 & 256-128-64 \\
Number of epochs            & 150 & 300  & 300 \\
Scaled Similarities         & \cmark & \cmark & \cmark \\
Scaled Input                & \cmark & \cmark & \cmark \\
Number of Prototypes        & 1 & 1 & 1 \\
Loss Type                   & Smooth margin & Smooth margin & Smooth margin \\
Activation Type             & ReLU & ReLU & ReLU \\
\bottomrule
\end{tabular}}
\end{table}

\begin{table}[ht]
\centering
\caption{Detailed layer configurations for AlexNet, VGG11, and VGG16 used in HFF training. All deep models do not scale similarities or the input. }
\label{tab:model_archs}
\resizebox{0.7\linewidth}{!}{
\begin{tabular}{lccccccc}
\toprule
\textbf{Model} & \textbf{Layer} & \textbf{Type} & \textbf{Out Ch.} & \textbf{Kernel} & \textbf{Stride} & \textbf{Pad} & \textbf{Aux Ch.}  \\
\midrule
\multirow{8}{*}{\textbf{AlexNet}} 
& L0  & Conv & 64  & 11 & 4 & 2 & 256   \\
& L1  & Conv & 192 & 5  & 1 & 2 & 128   \\
& L2  & Conv & 384 & 3  & 1 & 1 & 64  \\
& L3  & Conv & 256 & 3  & 1 & 1 & 32   \\
& L4  & Conv & 256 & 3  & 1 & 1 & 32   \\
& L5  & FC   & 4096 & -- & -- & -- & --   \\
& L6  & FC   & 4096 & -- & -- & -- & --  \\
& L7  & FC   & 1000 & -- & -- & -- & --  \\
\midrule

\multirow{11}{*}{\textbf{VGG11}} 
& L0  & Conv & 64  & 3 & 1 & 1 & 512  \\
& L1  & Conv & 128 & 3 & 1 & 1 & 512   \\
& L2  & Conv & 256 & 3 & 1 & 1 & 512   \\
& L3  & Conv & 256 & 3 & 1 & 1 & 512   \\
& L4  & Conv & 512 & 3 & 1 & 1 & 256   \\
& L5  & Conv & 512 & 3 & 1 & 1 & 256   \\
& L6  & Conv & 512 & 3 & 1 & 1 & 128   \\
& L7  & Conv & 512 & 3 & 1 & 1 & 64   \\
& L8  & FC   & 512 & -- & -- & -- & --  \\
& L9  & FC   & 512 & -- & -- & -- & --   \\
& L10 & FC   & num\_classes & -- & -- & -- & --   \\
\midrule
\multirow{16}{*}{\textbf{VGG16}} 
& L0  & Conv & 64  & 3 & 1 & 1 & 512   \\
& L1  & Conv & 64  & 3 & 1 & 1 & 512   \\
& L2  & Conv & 128 & 3 & 1 & 1 & 512   \\
& L3  & Conv & 128 & 3 & 1 & 1 & 512   \\
& L4  & Conv & 256 & 3 & 1 & 1 & 512  \\
& L5  & Conv & 256 & 3 & 1 & 1 & 512   \\
& L6  & Conv & 256 & 3 & 1 & 1 & 256   \\
& L7  & Conv & 512 & 3 & 1 & 1 & 256   \\
& L8  & Conv & 512 & 3 & 1 & 1 & 256   \\
& L9  & Conv & 512 & 3 & 1 & 1 & 128   \\
& L10 & Conv & 512 & 3 & 1 & 1 & 128   \\
& L11 & Conv & 512 & 3 & 1 & 1 & 64   \\
& L12 & Conv & 512 & 3 & 1 & 1 & 64   \\
& L13 & FC   & 4096 & -- & -- & -- & --   \\
& L14 & FC   & 4096 & -- & -- & -- & --   \\
& L15 & FC   & num\_classes & -- & -- & -- & --   \\
\bottomrule
\end{tabular}}
\end{table}

\subsubsection*{Effect of Hyperparameters in our Loss Function}
In \autoref{tab:special_cases_tau}, we show how the the number of prototypes $P$ and temperature $\tau$ change the layerwise objective function. It is interesting to note that our loss includes CE as a special case when $P=1$ and $\tau=1$. Empirical results show that the CE loss underperforms in the hypersphere. While the accuracy drops for the extreme $\tau$ values of 0.001 and 1000, our model's performance is stable in the finite range.
\begin{table}[ht]
\vspace{-2mm}
\centering
\caption{Special cases of our loss \& $\tau$ sensitivity on C100 (CNN).}
\label{tab:special_cases_tau}
\small\resizebox{0.6\linewidth}{!}{
\begin{tabular}{cccclc}
\toprule
$\textbf{P}$ & \textbf{$\tau$ Range} & \textbf{$\tau$ Val} & \textbf{Regime} & \textbf{Equivalent Method} & \textbf{Acc} \\
\midrule

\multirow{7}{*}{$1$}
 & $\rightarrow0$ & $0.001$ & Mean & Center Loss Var. &  25.04\\
 \cmidrule(lr){3-6}
 & $=1$ & $1$ & Softmax & CE & 48.74 \\
 \cmidrule(lr){3-6}
 & \multirow{4}{*}{Finite}
 & $5$  & \multirow{4}{*}{Scaled Softmax} & \multirow{4}{*}{Proxy-NCA (Ours) } &  52.46\\
 &  & $10$ &  &   &  \textbf{54.54}\\
 &  & $20$ &  &   & 53.63 \\
 \cmidrule(lr){3-6}
 & $\rightarrow\infty$ & $1000$ & Hard Max & Hard Neg Mining &  30.21\\
\midrule

\multirow{2}{*}{$>1$}
 & \multirow{2}{*}{Finite}
 & $5$  & \multirow{2}{*}{Soft Proto Mix} & \multirow{2}{*}{Ours} &  50.78\\
 &  & $10$ &  &  & 53.63 \\

\bottomrule
\end{tabular}}
\end{table}

\section{Additional Experiments}
\subsection{Analysis on the Number of Prototypes} 
\label{sec:prototype_analysis} 
A key component of our HFF model is the set of learnable prototypes. 
To understand the impact of the number of prototypes per class, $P$, we conducted an ablation study, with results presented in \autoref{tab:prototypes_ablation}. 
Our findings indicate that the optimal value of $P$ is a function of dataset complexity, reflecting a trade-off between model capacity and the risk of overfitting. 

\textbf{MNIST.} For the relatively simple MNIST dataset, where intra-class variation is low, a single prototype ($P=1$) per class is sufficient to achieve the close to the highest accuracy. 
The classes are largely unimodal, and adding more prototypes does not provide a significant benefit ($\le$0.04 p.p increase). 

\textbf{CIFAR.} On the more complex CIFAR-10 dataset, we see varying trends across MLP and CNN variants. With HFF-MLP, the accuracy consistently degrades as $P$ increases, suggesting that a single prototype adequately captures the class structure. Whereas HFF-CNN shows imporvement of 0.4 p.p with four prototypes as compared to one prototype. 
This suggests that the visual classes in CIFAR-10 exhibit some multi-modality (\emph{e.g.}, a "dog" class may contain distinct clusters for different breeds or resemble another class like "cats") that can be captured more effectively by other prototypes. 
However, as $P$ increases beyond four, accuracy begins to decline, indicating that the model starts to overfit with too many parameters dedicated to each class manifold. 

\textbf{Takeaway.} For our chosen datasets, a single, robust prototype ($P=1$) per class provides the best generalization. 
For simplicity and efficiency, we therefore adopt a single prototype per class in all experiments. Nonetheless, more complex or multi-modal datasets may benefit from increasing $P$ adaptively, highlighting an interesting direction for future work.

\begin{table}[ht]
    \centering
    \label{tab:prototypes_ablation}
    \caption{Ablation on the number of prototypes per class ($P$).
    We evaluate HFF-MLP and HFF-CNN with varying numbers of class-specific prototypes ($P$). Bold denotes the best accuracy for each setting, and gray cells indicate the chosen configuration for all experiments. Our results show that increasing $P$ expands model capacity but also increases parameters.
}
   \resizebox{0.4\linewidth}{!}{%
    \begin{tabular}{l|l|r|c}
        \toprule
         \textbf{Dataset} & \textbf{Method} & \textbf{proto} & \textbf{Acc}  \\
         \midrule
         MNIST & \multirow{4}{*}{HFF MLP} & \textbf{1} & \cellcolor{gray!20}98.53 \\
         & & 2 & 98.47 \\
         & & 4 & 98.51 \\
         & & 8 & \textbf{98.55} \\
         \cmidrule{2-4}
         & \multirow{4}{*}{HFF CNN} & \textbf{1} & \cellcolor{gray!20}99.08 \\
         & & 2 &  99.07\\
         & & 4 &  99.11\\
         & & 8 &  \textbf{99.15}\\
         \midrule 
         CIFAR-10 & \multirow{4}{*}{HFF MLP} & 1 & \cellcolor{gray!20}\textbf{61.33} \\
         & & 2 & 60.38 \\
         & & 4 & 59.52 \\
         & & 8 & 58.89 \\
         \cmidrule{2-4}
         & \multirow{4}{*}{HFF CNN} & 1 & \cellcolor{gray!20}83.18 \\
         & & 2 &  82.94 \\
         & & 4 & \textbf{83.58} \\
         & & 8 & 83.15 \\
        \bottomrule 
    \end{tabular}
    }
    
\end{table}

\subsection{Explainability} 
\subsubsection*{Layerwise Accuracies}

A key advantage of locally trained networks is that every layer produces its own class prediction, enabling fine-grained interpretability and diagnostics. In HFF, we obtain these layerwise predictions using the argmax of each layer’s gscore vector.
Table \autoref{tab:layer_accs} reports the resulting accuracies across datasets. For CIFAR-10 and CIFAR-100, the fifth convolutional layer exhibits notably higher accuracy (83.82\%), suggesting that meaningful classification can occur before the final fully connected layer. In such cases, we find that deploying predictions with single or multiple intermediate layers is feasible and often effective. For ImageNet1k, accuracy continues to improve gradually toward the deeper layers, reflecting the higher complexity of the dataset.
\begin{table*}[ht]
\centering
\caption{Layerwise accuracies of HFF-VGG models. \textbf{Bold} marks the best, and \colorbox{gray!20}{gray} shows the trainable layers in transfer learning tasks.}
\label{tab:layer_accs}
\resizebox{\linewidth}{!}{%
\begin{tabular}{ccccccccccccccccc}
\toprule
\textbf{Dataset} & \textbf{L0} & \textbf{L1} & \textbf{L2} & \textbf{L3} & \textbf{L4} & \textbf{L5} & \textbf{L6} & \textbf{L7} & \textbf{L8} & \textbf{L9} & \textbf{L10} & \textbf{L11} & \textbf{L12} & \textbf{L13} & \textbf{L14} & \textbf{L15}\\
\midrule
C10 & 49.73 & 75.14 &  82.33 &  83.26 &  \textbf{83.82} &  83.74 &  83.79 &  83.81 &  83.82 &  83.79 &  83.48 & -& -& -& -& -\\
C100 &  19.55 & 40.34 & 52.53 & 53.41 & 54.25 & 54.27 & 54.57 & 54.39 & 54.43 & 54.49 & \textbf{54.66} & -& -& -& -& -\\
IN1k &  3.32 & 6.51 & 10.38 & 11.34 & 14.22 & 16.02 & 15.82 & 19.12 & 20.93 & 21.20 & 22.28 & 22.35 & 23.08 & 24.44 & 25.52 & \textbf{25.66}\\
\midrule
\multicolumn{17}{l}{\textit{Transfer Learning (from ImNet1k)}} \\
IN1k-\textit{freeze} &  0.12 & 0.13 & 0.07 & 0.07 & 0.10 & 0.14 & 0.06 & 0.10 & 0.13 & 0.12 & 0.14 & 0.08 & 0.12 & \cellcolor{gray!20}60.83 & \cellcolor{gray!20}65.10 & \cellcolor{gray!20}\textbf{65.95}\\
IN1k-\textit{train} &  \cellcolor{gray!20}2.56 & \cellcolor{gray!20}5.77 & \cellcolor{gray!20}10.01 & \cellcolor{gray!20}10.99 & \cellcolor{gray!20}13.95 & \cellcolor{gray!20}15.85 & \cellcolor{gray!20}15.52 & \cellcolor{gray!20}19.01 & \cellcolor{gray!20}20.78 & \cellcolor{gray!20}20.66 & \cellcolor{gray!20}21.66 & \cellcolor{gray!20}21.73 & \cellcolor{gray!20}22.41 & \cellcolor{gray!20}24.73 & \cellcolor{gray!20}26.30 & \cellcolor{gray!20}\textbf{26.74} \\
C100-\textit{freeze} &  1.07 & 1.33 & 0.57 & 1.30 & 0.61 & 1.78 & 0.68 & 1.20 & \cellcolor{gray!20}40.45 & \cellcolor{gray!20}\textbf{42.82} & \cellcolor{gray!20}42.68 & -& -& -& -& -\\
C100-\textit{train} &  \cellcolor{gray!20}12.52 & \cellcolor{gray!20}31.24 & \cellcolor{gray!20}48.43 & \cellcolor{gray!20}55.09 & \cellcolor{gray!20}57.79 & \cellcolor{gray!20}\textbf{58.57} & \cellcolor{gray!20}58.40 & \cellcolor{gray!20}58.28 & \cellcolor{gray!20}58.38 & \cellcolor{gray!20}58.49 & \cellcolor{gray!20}58.39 & -& -& -& -& -\\
C10-\textit{freeze} &  5.93 & 10.86 & 10.2 & 9.31 & 10.34 & 8.77 & 7.76 & 10.81 & \cellcolor{gray!20}67.22 & \cellcolor{gray!20}\textbf{68.28} & \cellcolor{gray!20}67.28 & -& -& -& -& -\\
C10-\textit{train} &  \cellcolor{gray!20}46.31 & \cellcolor{gray!20}71.72 & \cellcolor{gray!20}83.9 & \cellcolor{gray!20}85.08 & \cellcolor{gray!20}85.78 & \cellcolor{gray!20}85.91 & \cellcolor{gray!20}85.91 & \cellcolor{gray!20}\textbf{86.01} & \cellcolor{gray!20}85.99 & \cellcolor{gray!20}85.99 & \cellcolor{gray!20}85.83 & -& -& -& -& -\\
\bottomrule
\end{tabular}%
}
\end{table*}

\textbf{Transfer Learning}
We further analyze layerwise behavior under transfer learning. We initialize the convolutional layers of our HFF-VGG models using ImageNet-pretrained VGG (VGG11 for CIFAR and VGG16 for ImageNet) and compare two regimes:
1. Transfer,\textit{freeze}: frozen convolutional layers with only final HFF FC layers trained; and 2. Transfer,\textit{train}: initialize from pretrained weights but allowing all layers and prototypes to adapt.

In Transfer,\textit{freeze} experiments, frozen convolutional backbones yield low early-layer accuracies with a sharp rise at the trained FC layers. Although ImageNet-transferred models achieve higher final accuracy (65.96\%), their interpretability remains limited. Moreover, when transferring between datasets (\emph{e.g.}, from Imagenet to CIFAR-100), the overall accuracy drops to 42.68\% from 54.66\%. We therefore relax the frozen-layer constraint, allowing all layers to fine-tune. This approach transfers IN1k representational knowledge while maintaining interpretability, resulting in improved performance(58.39\%), and more consistent layer-wise performance in fewer epochs than training from scratch (18 epochs vs 240 for CIFAR-10).

\subsubsection*{Activation Maps}
\begin{figure*}[ht]
  \centering
  \includegraphics[width=\linewidth]{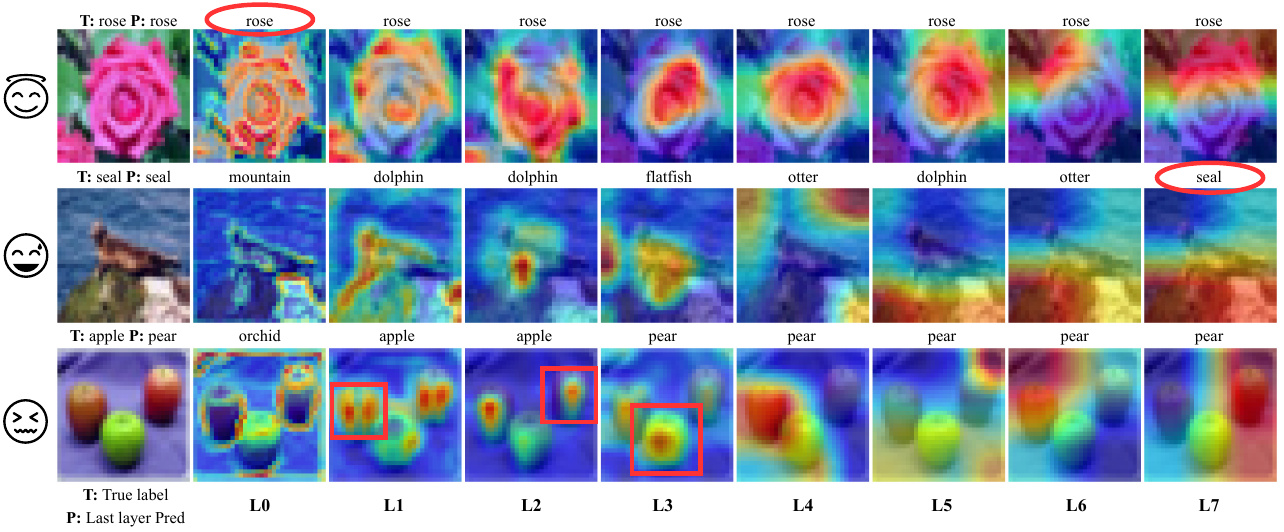}
  \caption{\textbf{Layerwise Activation Maps and Predictions in transfer-C100-\textit{train}.} \textit{Row 1:} An easy sample that was predicted correctly in all layers. \textit{Row 2:} The model predicted wrong but similar-looking aquatic animals in the early layers before getting the correct label. \textit{Row 3:} Shows a fault case, but we can interpret the reasoning. The model correctly labeled the red apples but mistook the green apple(L3) for a pear.}
  \label{fig:layerwise_cam}
\end{figure*}
In addition to quantitative metrics, activation maps, shown in \autoref{fig:layerwise_cam}, provide rich insights into the internal layers by highlighting the regions that drive each decision. These visualizations reveal easy vs. hard samples (rows 1 and 2), correct attention localization, dataset bias, and fault cases analysis (row 3), offering a clearer view of the network's reasoning behavior.

\subsection{Prototype UMAP}
\begin{figure*}[t]
  \centering
  \includegraphics[width=\linewidth]{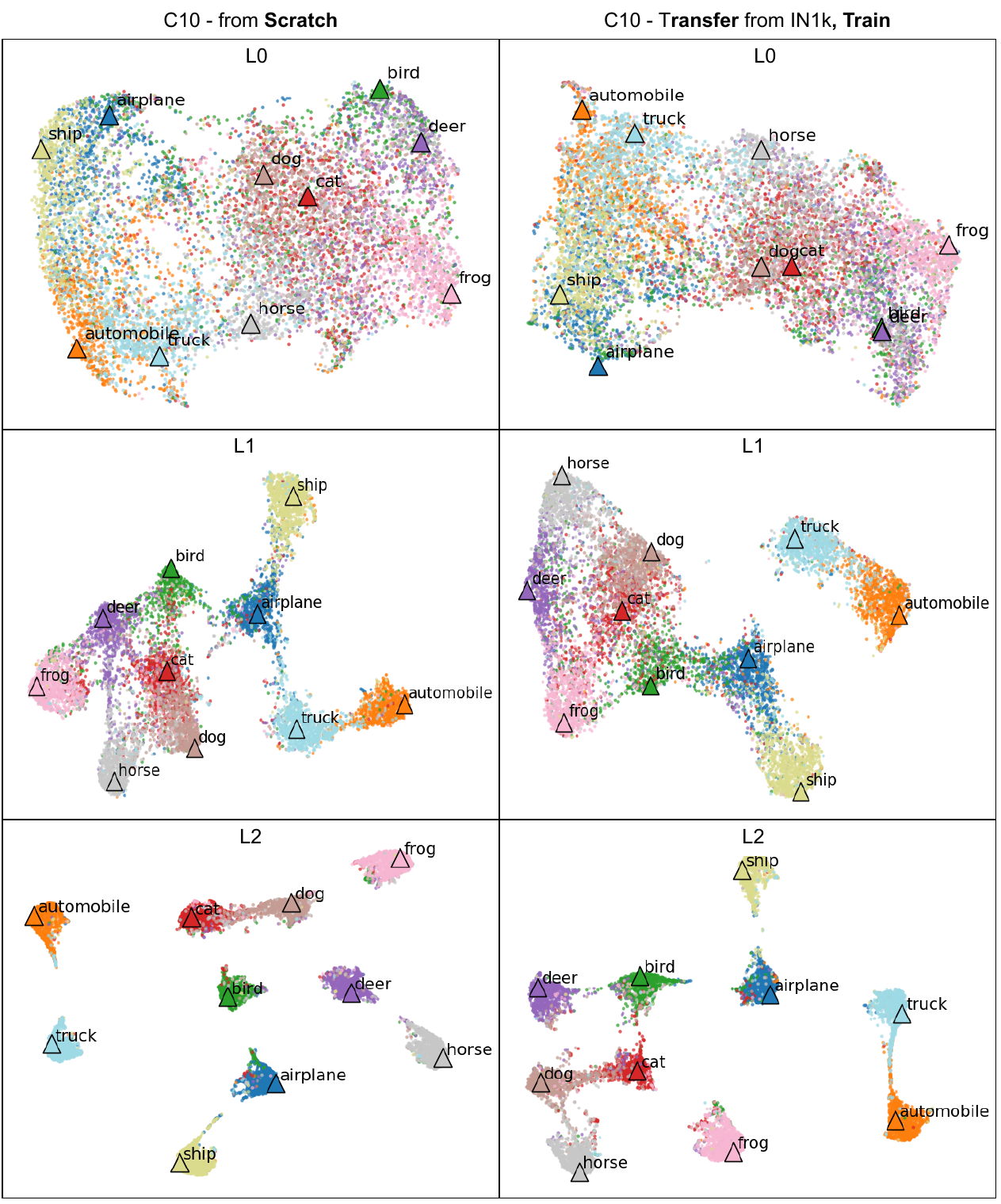}
  \caption{\textbf{UMAP of learned prototypes of C10}}
  \label{fig:fig_proto_umap_c10_1}
\end{figure*}

\begin{figure*}[ht]
  \centering
  \includegraphics[width=\linewidth]{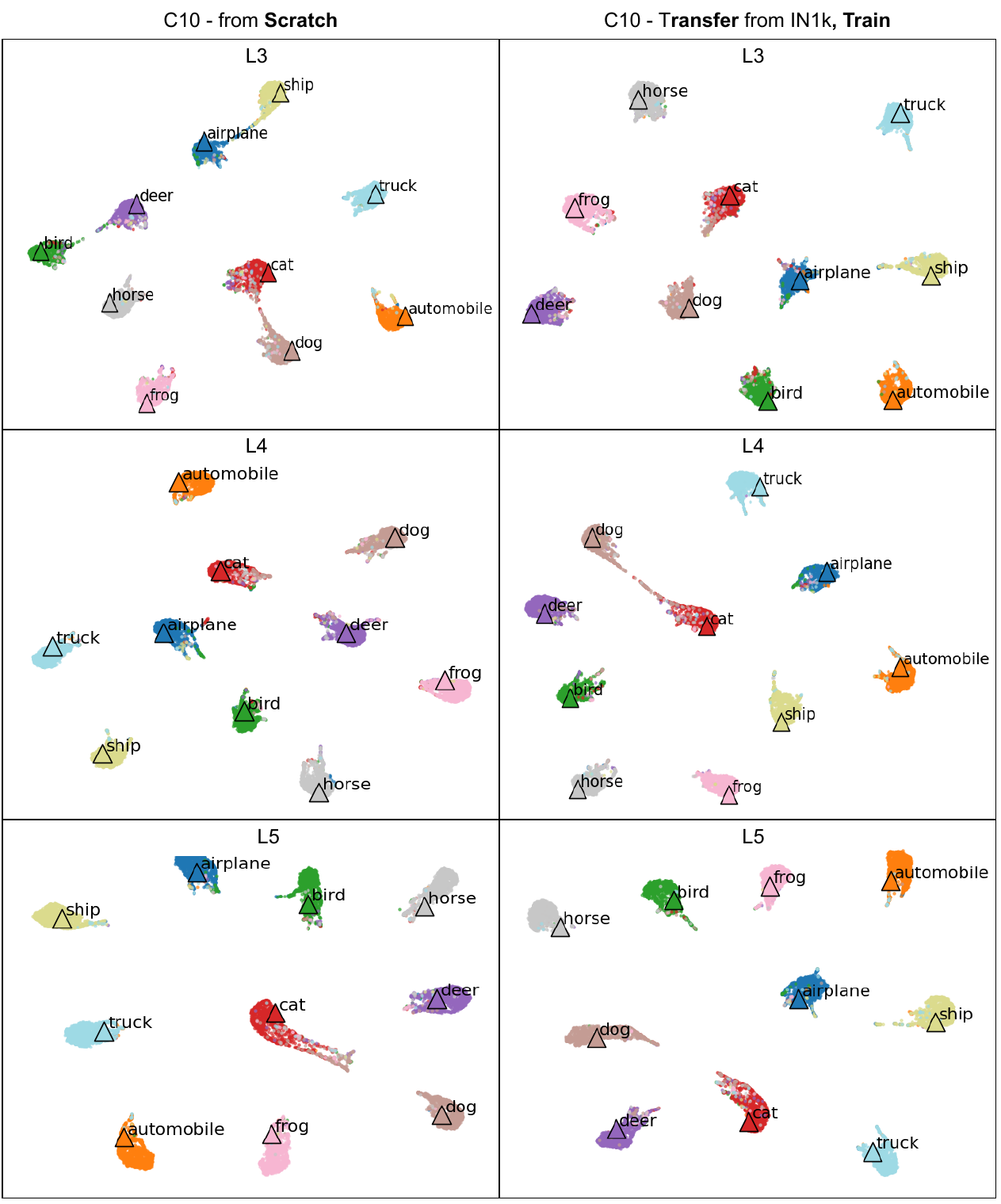}
  \caption{\textbf{UMAP of learned prototypes of C10}}
  \label{fig:fig_proto_umap_c10_2}
\end{figure*}

\begin{figure*}[ht]
  \centering
  \includegraphics[width=\linewidth]{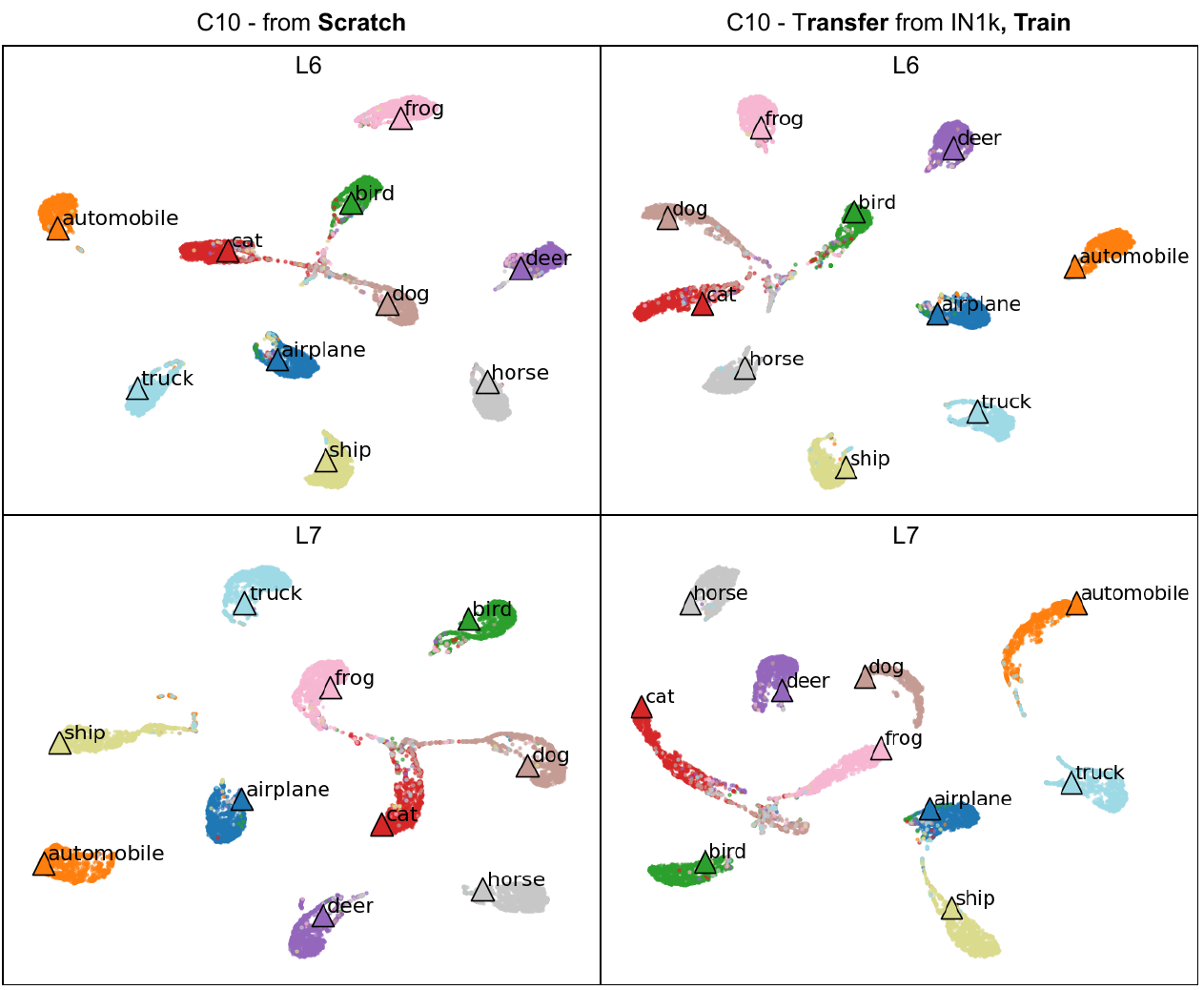}
  \caption{\textbf{UMAP of learned prototypes of C10}}
  \label{fig:fig_proto_umap_c10_3}
\end{figure*}

To better understand the geometry induced by HFF training, we visualize the learned class prototypes together with the corresponding feature clouds using UMAP. For each layer, we project both the unit-norm prototypes and a random subset of training features into a 2D manifold. These plots (as shown in \autoref{fig:fig_proto_umap_c10_1}, \autoref{fig:fig_proto_umap_c10_2}) and \autoref{fig:fig_proto_umap_c10_3}) reveal how early layers form loose, overlapping clusters, while deeper layers progressively contract each class into a tight region anchored around its prototype. The prototypes typically occupy stable, well-separated positions on the hypersphere, and the feature clouds align around them as training progresses, illustrating how the local objective shapes discriminative structure layer by layer. This visualization also exposes class confusion patterns and dataset bias, offering an interpretable view of how representations evolve within locally trained networks.

\label{sec:appendix_proto_umap}
\newpage
\newpage



\end{document}